%% file: neurips_2026.tex
\documentclass{article}

    \PassOptionsToPackage{numbers, compress}{natbib}
\usepackage[preprint]{neurips_2026}

\usepackage[utf8]{inputenc} % allow utf-8 input
\usepackage[T1]{fontenc}    % use 8-bit T1 fonts
\usepackage{xurl}
\usepackage{url}            % simple URL typesetting
\usepackage{hyperref}       % hyperlinks

\usepackage{booktabs}       % professional-quality tables
\usepackage{amsfonts}       % blackboard math symbols
\usepackage{nicefrac}       % compact symbols for 1/2, etc.
\usepackage{microtype}      % microtypography
\usepackage{xcolor}         % colors
\usepackage{enumitem}
\usepackage{tcolorbox}
\usepackage{tikz}
\usepackage{twemojis}
\usepackage{graphicx} 
\usepackage{multicol}
\usepackage{pdfpages}
\usepackage{float}
\usepackage{amsmath}

\definecolor{darkorchid}{rgb}{0.6, 0.2, 0.8}

\definecolor{GREEN}{HTML}{B4E0B0}
\definecolor{BLUE}{HTML}{8EC6DF}
\definecolor{PURPLE}{HTML}{D9B4FF}
\definecolor{GRAY}{HTML}{E6E6E6}
\definecolor{LIGHT}{HTML}{FAFAFA}

\definecolor{RED}{HTML}{F4CCCC}        % Soft red/warning
\definecolor{ORANGE}{HTML}{FFD966}     % Warm orange/highlight
\definecolor{YELLOW}{HTML}{FFE599}     % Soft yellow/note
\definecolor{TEAL}{HTML}{B2DFDB}       % Fresh teal/data
\definecolor{PINK}{HTML}{F0B5F0}       % Soft pink/people
\definecolor{MINT}{HTML}{C8E6C9}       % Mint green/success
\definecolor{LAVENDER}{HTML}{E1D5E7}   % Light purple/creative
\definecolor{PEACH}{HTML}{F2B88C}      % Peach/warmth
\definecolor{MUSTARD}{HTML}{F4E0A8}    % Mustard/caution
\definecolor{CORAL}{HTML}{F4B9B9}      % Coral/attention
\definecolor{SAGE}{HTML}{B5C9A8}       % Sage green/nature
\definecolor{NAVY}{HTML}{B3C7E3}       % Soft navy/professional
\newcounter{globalcounter}
\newcommand{\numitem}[1]{%
  \par\noindent
  \refstepcounter{globalcounter}%   ← IMPORTANT FIX
  \hspace{1.7em}%
  \hangindent=3.8em%
  \hangafter=1%
  \makebox[1.5em][l]{\arabic{globalcounter}.}%
  \hspace{0em}%
  #1%
  \par%
  \vspace{0.5pt}%
}

\newcommand{\mybox}[2]{%
    \colorbox{#1}{#2}%
}

\newcommand{\resetcounter}{\setcounter{globalcounter}{0}}

\newcommand{\opt}[1]{$\circ$\,#1\hspace{0.1em}}

\newcommand{\subtitle}[1]{%
  \vspace{1pt}\textbf{#1}\par\vspace{2pt}
}

\newenvironment{attributes}[1][]{
  \begin{itemize}[
   label=\textbullet,
    leftmargin=3.7em,    
    itemsep=0pt,
    topsep=0pt,
    #1
  ]
}
{
  \end{itemize}
}

\title{LLJ Cards: \\Best practices for the Use of LLMs as Judges}

\author{%
  Khaoula Chehbouni$^{1,2}$ \quad
  Melina Medjdoub$^{3}$ \quad
  Florian Carichon$^{1,2}$ \AND
  Golnoosh Farnadi$^*$ $^{1,2}$ \quad
  Jackie Chi Kit Cheung$^*$$^{1,2}$ \\
  $^1$McGill University \\
  $^2$Mila - Quebec AI Institute \\
  $^3$Concordia University \\
  \texttt{khaoula.chehbouni@mila.quebec}\\ 
}

\begin{document}

\def\thefootnote{*}\footnotetext{These authors contributed equally to the work}
\maketitle

\begin{abstract}
In recent years, large language models (LLMs) have emerged as a popular alternative for evaluation. Often referred to as LLMs as judges~(LLJs), these systems have been widely adopted by researchers and practitioners across a broad range of measurement tasks, driven by their strong performance, scalability, and cost-effectiveness relative to human judgment. However, a growing body of work has shown that the use of LLJs raise concerns about their validity and reliability as evaluators. Existing efforts to address these challenges have largely focused on developing bias-mitigation techniques and refining prompting strategies. While these approaches represent an important step forward, they primarily offer technical fixes and leave a more fundamental challenge unaddressed: the lack of standardized, transparent, and reproducible evaluation practices.
In this paper, we introduce \textbf{LLJ Cards}, a framework that synthesizes best practices from measurement theory, natural language generation, and machine learning literature into practical guidelines for LLJ-based evaluations. While LLJs offer a promising path toward scalable evaluation, their effective use requires grounding in rigorous evaluation principles to ensure validity, reliability, and reproducibility. \textbf{LLJ Cards} addresses this need by providing a structured framework for applying these principles in the design and reporting of automated evaluations.
% , transparency, and trustworthiness.
\end{abstract}

\section{Introduction}
\label{sec:introduction}

In recent years, large language models (LLMs) as judges~(LLJs) have gained significant popularity among machine learning~(ML) and natural language processing~(NLP) researchers and practitioners~\citep{gu_survey_2025, li_llmsasjudges_2024}. Their use has also expanded well beyond computer science, reaching into the social sciences~\citep{savelka_unlocking_2023, heseltine_large_2024}. This uptake is largely driven by their strong alignment with human judgments, which have traditionally served as the gold standard in measurement, as well as their perceived effectiveness as evaluators~\citep{liu_geval_2023, bai_constitutional_2022}. In addition, their flexibility, scalability, and cost efficiency have contributed to their adoption and integration into standard evaluation practices~\citep{chehbouni_neither_2025}. 
Despite this widespread enthusiasm, a body of work has cautioned that such adoption may be premature, as numerous issues related to the validity and reliability of LLJs have been documented~\citep{zheng_large_2023, shi_large_2023, si_large_2024, hu_are_2024, feuer_style_2024, li_llms_2025, baumann_large_2025, guerdan_validating_2025}.

Human evaluation, despite its drawbacks in practice~\citep{clark_all_2021, zhou_deconstructing_2022},  is supported by an established body of literature that articulates norms to ground evaluations in their broader context~\citep{sabou_corpus_2014, feng_acquiring_2009}. In contrast, LLJs lower the barrier to evaluation to an unprecedented degree. This offers scalability and ease of use, but may encourage decontextualized or under-specified evaluation setups~\citep{chehbouni_neither_2025, dietz_principles_2025, guerdan_validating_2025}. 
\citet{crockett_ai_2026} argue that many evaluations have become \textit{DEAD}---that is, \textit{decontextualized, engineered, anonymized, and disembodied}---a characterization that maps closely onto current uses of LLJs. Decontextualized use of LLJs refers to their deployment with little specification of the intended audience for the evaluation or the domain of application. For instance, work leveraging LLJs for text summarization~\citep{liu_geval_2023, fu_gptscore_2024, song_finesure_2024} rarely clarifies what types of summaries are being evaluated or for which real-world applications. Evaluations are also highly \textit{engineered}: in safety assessment, complex sociotechnical notions of harm are often reduced to simplified judgments, such as binary preferences over whether content is harmful or not~\citep{chehbouni_safety_2025}. Finally, many evaluation setups remain \textit{anonymized}. A striking example is data labeling, where annotator demographics are often omitted, or, in the case of LLJs, where the positionality of the model is left unspecified~\citep{cambo_model_2022}, despite extensive evidence that such factors meaningfully shape the construction of datasets and evaluation outcomes~\citep{scheuerman_datasets_2021}~\footnote{As \citet{crockett_ai_2026}'s work was initially targeted towards cognitive scientist, embodied intelligence is outside the scope of this paper.}. 
As a result, LLJs run the risk of exacerbating known problems in evaluation by reducing the ``good friction'' that is rigorous operationalization and systematic validation.
While researchers have proposed methods to test these models~\citep{li_split_2024, xie_sorrybench_2024}, improve their reliability~\citep{schroeder_can_2025, dietz_principles_2025}, or automatically validate their use~\citep{ghiasvandmohammadkhani_checklist_2025, guerdan_validating_2025}, existing solutions mostly focus on targeted technical fixes. In this paper, we examine how practitioners and researchers evaluation practices can be improved.

Documentation is recognized as an important mechanism for improving both research and development practices. \citet{benderDataStatementsNatural2018} argue that documenting design decisions can advance both scientific and ethical practice by encouraging critical reflection on how these decisions shape system behavior, thereby supporting generalizability and reproducibility. Documentation can also promote more responsible AI development by disclosing potential limitations and shortcomings and supporting external auditing~\citep{chmielinski_clear_2024}. In parallel, research on reproducibility has shown that thorough documentation can facilitate system reuse while reducing the risk of misinterpreting or overstating system capabilities~\citep{magnusson_reproducibility_2023}. 
Building on these insights, we introduce \textbf{LLJ Cards}, a framework for improving the design and validation of LLJ-based evaluations while increasing transparency. 
Our framework draws on established practices from measurement theory~\citep{adcock_measurement_2001, jacobs_measurement_2021, xiao_evaluating_2023}, the computational social science~\citep{vanderlee_best_2019, sabou_corpus_2014}, natural language generation evaluation (NLG)~\citep{zhou_deconstructing_2022, howcroft_twenty_2020, alexandrova_adequacy_2010} and responsible AI research~\citep{selbst_fairness_2019, blodgett_stereotyping_2021}. 
This framework supports practitioners in designing and validating LLMs as judges through a structured set of guiding questions that promote more rigorous, reproducible, and transparent evaluation practices. Our cards make explicit the design choices underlying an evaluation and situate its results within the broader evaluation context, enabling clearer reporting of system capabilities, facilitating reuse, and helping clarify the extent to which findings may generalize beyond the original evaluation setting. By prompting practitioners to systematically reflect on their design choices, \textbf{LLJ Cards} provides a structured basis for transparent reporting and validation.

\section{Related Work}
\label{sec:background}
% In this section, we situate our work across three main areas: LLJs, measurement theory, and the literature on documenting ML and NLP practices.

\noindent\textbf{LLMs as Judges.}
In recent years, LLJs have emerged as a preferred evaluation method across a wide range of tasks and applications~\citep{bai_constitutional_2022, liu_geval_2023, kocmi_large_2023, wang_chatgpt_2023, huang_chatgpt_2023, fu_gptscore_2024, lee_rlaif_2024,chehbouni_neither_2025}.
% , driven by early findings highlighting their strong correlation with human judgment, their effectiveness as evaluators, and their scalability and cost-efficiency~\citep{bai_constitutional_2022, liu_geval_2023, kocmi_large_2023, wang_chatgpt_2023, huang_chatgpt_2023, fu_gptscore_2024, lee_rlaif_2024,chehbouni_neither_2025}.
% Indeed, they can assess outputs across a range of criteria while also generating explanations to justify their evaluations or provide feedback when needed, reducing reliance on costly human evaluation or highly specialized metrics. To better reflect the broad range of potential uses of LLJs,
% \citet{li_llmsasjudges_2024} introduce a structured framework for integrating different evaluation paradigms using LLJs by formalizing the evaluation process as an evaluation function that takes as input an \textit{evaluation type}, an \textit{evaluation criterion}, an \textit{evaluation item}, and optional \textit{references}. This function returns an \textit{evaluation result}, and may additionally produce an \textit{explanation} and \textit{feedback} when specified.
% This evaluation function can be categorized into three configurations~\citep{li_llmsasjudges_2024}: a single LLJ, multiple LLJs (usually referred to as a jury) or an LLJ with a human-in-the-loop. 
% We refer the reader to their survey for additional details~\citep{li_llmsasjudges_2024}. 
While LLJs offer an important opportunity to enable scalable evaluation, 
% ---a pressing challenge with the increasing adoption of generative models---
a growing body of work has shown potential limitations in their use, highlighting issues of reliability~\citep{zheng_large_2023,shi_large_2023, li_llms_2025, baumann_large_2025}, validity~\citep{si_large_2024, hu_are_2024, feuer_style_2024, guerdan_validating_2025}, and robustness~\citep{raina_llmasajudge_2024}, as well as the extent to which LLJ-based evaluation processes may embed or amplify biases~\citep{ye_justice_2024, wang_large_2024, panickssery_llm_2024, liu_llms_2024, li_calibraeval_2024, koo_benchmarking_2024, chen_humans_2024, li_preference_2025}. 
\citep{chehbouni_neither_2025} review the literature on the topic and link some of these pitfalls to existing evaluation practices in the field.
Furthermore, prior work addressing concerns about LLJs validity and reliability has proposed benchmarks and automated test suites to evaluate their performance~\citep{dietz_principles_2025, feng_are_2025, ghiasvandmohammadkhani_checklist_2025}. For instance, \citep{dietz_principles_2025} identify common pitfalls in the use of LLJs and outline experimental protocols to mitigate them. \citep{feng_are_2025} introduce an automated evaluation suite for assessing LLJ quality, while \citep{ghiasvandmohammadkhani_checklist_2025} propose a training-free framework that generates checklists to improve LLJ systems in multilingual settings.
In contrast to this body of work focused on technical interventions, \textbf{LLJ Cards} aim to improve practitioners' evaluation practices.

\begin{figure}[H]
    \centering
    \includegraphics[width=\textwidth]{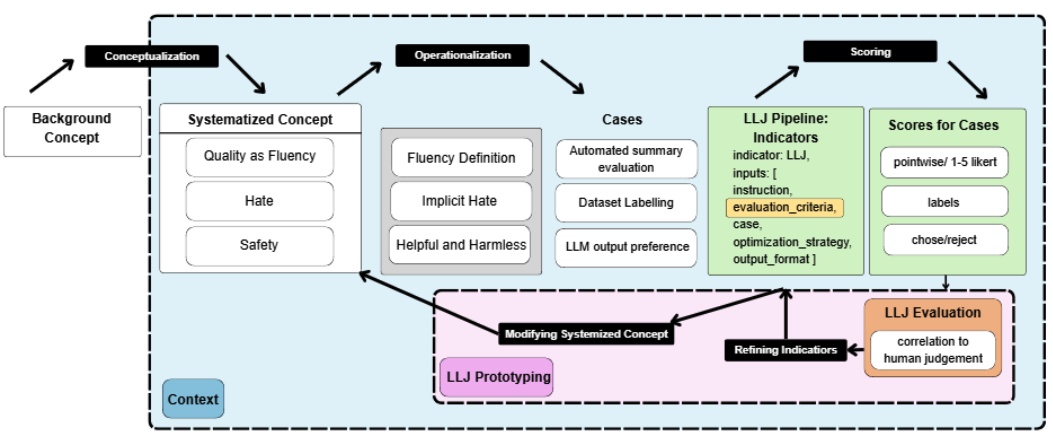}
    \caption{This figure presents an overview of \citep{adcock_measurement_2001} framework applied to LLJ-based evaluation.}
    \label{fig:measurement_example}
\end{figure}

\noindent\textbf{Measurement Theory.}
% Measurement theory offers a conceptual foundation for formalizing and evaluating the validity and reliability of evaluation methods~\citep{}. 
Originating in the quantitative social sciences, measurement theory has been shaped by longstanding efforts to develop rigorous approaches for operationalizing and validating theoretical, often abstract, constructs (e.g., democracy).
\citet{adcock_measurement_2001} introduce a four-level framework to depict the relationship between concepts and observations: (1) the background concept, which includes all possible definitions of the concept to be measured; (2) the systematized concept, which is a specific formulation of that concept along with its definition; (3) the indicators, which are the measures used to score cases; and (4) the scores, which are the resulting outcomes of the measurement process. They define measurement as encompassing everything required to move from a systematized concept to observed scores~\citep{adcock_measurement_2001}: within this framework, a measurement is valid when the scores produced by an indicator---in our case an LLM as judge---can be meaningfully interpreted in relation to the systematized concept it is intended to assess. A body of work has looked at how this framework could be adapted for NLG evaluation~\citep{zhou_deconstructing_2022, jacobs_measurement_2021, wallach_evaluating_2024}, we build on top of this literature and structure our \textbf{LLJ Cards} around it~\footnote{Consequently, we do not include considerations related to the background concept.}.  See Figure~\ref{fig:measurement_example} for an illustration.

% in \S\ref{app:figure_measurement} illustrates how our framework maps onto an LLJ pipeline.
% \khaoula{Figure X present this framework applied to LLJs.}
% Conceptualization is the task of formulating a systematized concept based on the background concept, while operationalization encompasses developing indicators based on the systematized concept. 
% \citet{adcock_measurement_2001} define measurement as encompassing everything required to move from a systematized concept to observed scores: within this framework, a measurement is valid when the scores produced by an indicator---in our case an LLM as judge---can be meaningfully interpreted in relation to the systematized concept it is intended to assess. Any discrepancies between the two are referred to as measurement error. Such error takes two main forms. Systematic error introduces consistent bias and therefore undermines validity. Random error, which arises when repeated applications of the same measurement procedure produce inconsistent results, is instead associated with reliability.
% We structure 

\noindent\textbf{Documenting Practices.} 
Responsible AI research has long emphasized the importance of documenting technical artifacts such as datasets, models, and systems~\citep{benderDataStatementsNatural2018, mitchellModelCardsModel2019, arnoldFactSheetsIncreasingTrust2019, madaioCoDesigningChecklistsUnderstand2020, gebruDatasheetsDatasets2021, bhardwajMachineLearningData2024, liuECBDEvidenceCenteredBenchmark2024, zhuEstablishingBestPractices2025}. Among the most widely adopted proposals, \citet{gebruDatasheetsDatasets2021} introduce \textit{Datasheets for Datasets}, a structured set of guiding questions designed to document dataset creations, while \citet{mitchellModelCardsModel2019} extend this idea to model reporting through \textit{Model Cards}. These approaches aim to standardize documentation practices, support practitioners in recording key decisions about their artifacts, and improve transparency in the field. Similarly, \citet{liuECBDEvidenceCenteredBenchmark2024} focus on documenting practitioners’ design choices in system development and on guiding them in collecting validity evidence for benchmark-based measurement. While \citet{zhuEstablishingBestPractices2025} propose a checklist to guide practionners through agentic benchmarks assessment for validity.
We introduce \textbf{LLJ Cards} to guide LLJ users toward a more valid and reliable use of LLJs through a set of guiding questions grounded in evaluation best practices.

\section{Designing the LLJ Cards}
\label{sec:overview_framework}

% \khaoula{Ici on peut mettre quick summary + constructing the framework + idees behing = seamless et min efforts + comment c'est general + expected benefits + structure and how to use it (chronologie sur pilot study par e.g.)}

% maybe use this: Thus, we propose the \textbf{LLJ Cards} framework to support ML researchers and practitioners---hereafter referred to as ``LLJ users''---in developing valid and reliable measurement tools, thereby enabling them to fully realize the potential of LLJs for evaluation. To ensure that our framework is broadly applicable, we ground our approach in representative use cases for each of each of the three functionalities described by \citet{li_llmsasjudges_2024}, following \citep{chehbouni_representational_2024}: text summarization, safety alignment, and data labeling.

We introduce \textbf{LLJ Cards}, a framework to support ML researchers and practitioners---hereafter referred to as ``LLJ users''---in developing valid and reliable measurement tools.
% , thereby enabling them to fully realize the potential of LLJs for evaluation. 
\textbf{LLJ Cards} guide users through key considerations when designing and using LLJs, helping them avoid common pitfalls, adopt best practices, and promote more standardized and reproducible evaluation practices.
To ensure that our framework is broadly applicable, we ground our approach in representative use cases for each of each of the three functionalities described by \citet{li_llmsasjudges_2024}, following \citep{chehbouni_representational_2024}: text summarization, safety alignment, and data labeling.

%I put a compressed summary because i have too much method content that i moved into appendix
To design our \textbf{LLJ Cards} framework, we first identified the key  
% reviewed the literature on measurement theory~\citep{adcock_measurement_2001, messick_test_1980, messick_validity_1994}, as well as its applications in machine learning~\citep{jacobs_measurement_2021, wallach_evaluating_2024, xiao_evaluating_2023, chouldechova_shared_2024}, natural language generation eva~\citep{}, fairness~\citep{}, and related fields~\citep{}, 
% in order to identify the key 
desiderata of a successful evaluation~\citep{john_measurement_2000, mokkink_cosmin_2010, scholtes_what_2011, crockett_ai_2026, messick_test_1980, messick_validity_1987, messick_validity_1994, adcock_measurement_2001, Potter01081999}, before examining literature in each of the selected use cases.
% : text summarization, data labeling and safety alignment. 
For each use case, we review the literature to characterize what constitutes a ``good evaluation,'' document known limitations and failure modes, and extract best practices transferable to our setting.
We considered both automated and human-led evaluation methods, as well as relevant work from adjacent disciplines (e.g., information theory~\citep{wittrock_generation_1990, hidi_producing_1986, hill_writing_1991}, computational social science~\citep{prabhakaran_releasing_2021, díaz_crowdworksheets_2022, klie_analyzing_2024}, cybersecurity~\citep{peake_red_2021,wang_red_2025, purpura_building_2025, inie_summon_2025}), to complement our findings. 
We then reviewed LLJ applications for each use case to understand how they are employed in practice~\citep{liu_geval_2023, fu_gptscore_2024, chiang_can_2023, song_finesure_2024, wang_chatgpt_2023, liusie_llm_2024, törnberg_best_2024, lee_rlaif_2024, bai_constitutional_2022, sun_salmon_2023, inan_llama_2023}, alongside literature on their biases and limitations to identify current shortcomings~\citep{chehbouni_neither_2025, li_llms_2025, vallejovera_llms_2025}.
Based on this review, we translated the best practices and recommendations identified in the literature into the LLJ setting by designing a coherent workflow aligned with current LLJ practices.
% Appendix~\ref{ap:method_additional} provides additional details on our method for constructing the LLJ Cards. The remainder of this section provides an overview of our framework. 
% \khaoula{support ML researchers and practitioners---hereafter referred to as ``LLJ users''}

Our literature review led us to the following LLJ workflow: LLJ users first identify the cases they wish to evaluate and define the type of evaluation to conduct. They then select appropriate evaluation criteria and set up their LLJ pipeline. Next, they typically perform a form of LLJ tuning---such as model selection, prompt engineering, or hyperparameter tuning---before selecting their final pipeline. Finally, they use the LLJ pipeline to score all cases and report the results along with an overall assessment of the LLJ validity, typically by measuring its correlation with a reference metric of their choice.
However, this final validation step captures only one aspect of validity: whether the LLJ produces scores that align with a chosen reference metric. According to \citet{messick_validity_1994}, a valid evaluation should also consider the broader consequences of using the LLJ and its resulting scores.
% Additionally, validation should also encompass the assessment of the broader societal consequences of measurement~\citep{messick_validity_1994}.
% In particular, according to \citet{messick_validity_1994}, the consequential aspect of validity requires attention to both the potential consequences of the use of the evaluation  and the implications of using these scores as a basis for action.

Following this workflow, we introduce five cards to document and facilitate the use and understanding of LLJs: Context~(\ref{subsec:questions_context}), Evaluation Criteria~(\ref{subsec:questions_evaluation_criteria}), LLJ Prototyping~(\ref{subsec:questions_pilot}), LLJ Pipeline~(\ref{subsec:questions_llj_pipeline}), and LLJ Evaluation~(\ref{subsec:questions_llj_evaluation}). 

\newpage
\section{LLJ Cards: Best Practices for the Use of LLMs as Judges}
\label{sec:framework_lljs}
This framework is designed to help LLJ users throughout the process of using an LLJ. In this section, we first introduce a card before providing a more detailed description of recommended practices for LLJ users, along with examples illustrating their importance.

\input{llj_checklist_fixed}

\subsection{Context}
\label{subsec:questions_context}

In the \mybox{BLUE}{Context Card}, LLJ users are asked to provide contextual information across four axes: the application, the task, the cases, and the evaluation stakeholders.

 % For instance, when using an LLJ to assess the quality of generated text, it is important to reflect on the deployment setting of the model producing that text. Even in more theoretical work, grounding evaluations in plausible real-world applications remains essential to ensure responsible and meaningful use.

\noindent\textbf{\twemoji{1f30e} Application.} LLJ users should describe the real-world context of the application---or, when none is explicitly defined, outline plausible deployment scenarios (\textbf{Q.~\ref{q:app_domain}})---and discuss the known risks associated with that environment (\textbf{Q.~\ref{q:app_risks}}). For example, deploying an automated text summarization system in a healthcare setting---where summaries may inform decision-making---entails far higher stakes than using a similar system to generate book reviews on an online forum. Consequently, the criteria and priorities guiding the evaluation will differ accordingly.

\noindent\textbf{\twemoji{1f4dd} Task.} LLJ user should provide contextual information about the task in which the LLJ is being leveraged (\textbf{Q.\ref{q:task_what}}), its known limitations (\textbf{Q.\ref{q:task_limitations}}) and potential biases (\textbf{Q.\ref{q:task_bias}}). 
For example, safety classifiers and other automated content moderation tools are known to exhibit biases against marginalized groups, including Black~\citep{harris_honestly_2023} and queer~\citep{dorn_harmful_2024} online communities.

\noindent\textbf{\twemoji{1f4c1} Cases.} LLJ users should provide contextual information about the cases, since building an LLJ pipeline to evaluate them requires clearly defining what they are (e.g., generated automated summaries to evaluate, an existing dataset for labeling, etc.)~(\textbf{Q.~\ref{q:data_info}}). This includes how the data was collected, its domain, its scope and limitations, the values and perspectives encoded in it, and whether it is being reused for a purpose different from the one it was originally intended for. 
For example, user review ratings are often used as targets in sentiment-based opinion summarization to extract key product aspects~\citep{titov_modeling_2008}. However, because reviews often contain mixed opinions~\citep{hu_mining_2004}, ratings can miss such nuance, introducing noise and undermining evaluation when treated as ground truth~\citep{angelidis_summarizing_2018}. 

% Ratings have often been employed as a form of  supervision in sentiment-oriented opinion summarization, used to guide the extraction of salient product aspects \citep{titov_modeling_2008}. 
% However, reviews frequently exhibit mixed  opinions for instance, a generally negative review may still contain positive evaluations of specific attributes \citep{hu_mining_2004}. Consequently, treating ratings as ground-truth signals can introduce systematic noise, rendering evaluation and supervision based on ratings unreliable \citep{angelidis_summarizing_2018}.

\noindent\textbf{\twemoji{1f3e2} Stakeholders.} 
LLJ users should provide additional context about the evaluation process, including whether the LLJ is being used in a production setting or for an academic study, who is funding its use, and when the evaluation is taking place~(\textbf{Q.~\ref{q:llj_context}}). We also suggest discussing the intended audience, as it will influence the evaluation design (\textbf{Q.~\ref{q:llj_audience}}). 
Finally, we ask LLJ users to disclose their own background, as well as that of the broader research team~(\textbf{Q.~\ref{q:llj_positionality}}). This statement may include a discussion of how the LLJ users' educational background and training, as well as sociohistorical factors such as nationality, race, gender, class, or other identities, could influence the outcomes of the project~\citep{scheuerman_datasets_2021, Scheuerman2020HCIGF, raceforward_race_2015}. For example, \citet{patton_annotating_2019} emphasize the need for domain experts when annotating data from marginalized communities, showing strong disagreement between annotators familiar with gang-related activity and those who are not. Similarly, \citet{fleisig_when_2023} demonstrate that annotators’ demographics can influence labels in hate speech detection tasks.

The  \mybox{BLUE}{Context Card} aims to encourage LLJ users to ground their work within its broader sociotechnical context.
% We decompose this card into four categories, first, the broader context surrounding the technical artifact we want to evaluate (e.g. if it is an automated summarization system, where will it  be used?), information about the task itself (e.g., summarization), the cases being evaluated (e.g., summaries) and finally additional background 
\citet{grote_fairness_2024} defines the evaluation process as riddled with \textit{``epistemic and morally normative considerations''} as AI systems and other technical artifacts cannot be considered in isolation, but should instead be conceived as tools used for specific purposes with the potential to impact their societal environment~\citep{grote_fairness_2024}. This perspective aligns with scholars’ call for an \textit{adequacy-for-purpose view}~\citep{alexandrova_adequacy_2010, parker_model_2020}; that is, the importance of assessing models and other technical artifacts with respect to their success and reliability in a particular instance of use: they should stand with respect to a target, a user, a methodology, circumstances and a purpose~\citep{parker_model_2020}. This view also aligns with \citet{jones1999automatic}, who argue that a summary’s usefulness depends on \textit{who} it is for, \textit{what} it will be used for, and \textit{when} it will be used. Here, the technical artifact is seen as a solution in a problem space~\citep{parker_model_2020} that we want to define. We ask for additional background information about the evaluation stakeholders since prior work has shown that such information can meaningfully shape design choices. This reflects the importance of embracing reflexivity~\citep{scheuerman_datasets_2021} and acknowledging how our background and perspectives inform and constrain project results. We encourage LLJ users to not only think about their own background but also about the values and perspectives encoded in all the technical artifacts (e.g., datasets~\citep{scheuerman_datasets_2021} or models~\citep{cambo_model_2022}) they are using as measurement inherently involves value judgments at every stage of the process~\footnote{As discussed by \citet{heisenberg_physics_1958}: \textit{``What we observe is not nature itself, but nature exposed to our method of questioning''}.}.

\subsection{Evaluation Criteria}
\label{subsec:questions_evaluation_criteria}

In the \mybox{YELLOW}{Evaluation Criteria Card}, LLJ users should document their evaluation goals, including the systematized concept, how it is operationalized, and the limitations of that operationalization.

\noindent\textbf{\twemoji{1f4a1} Systematized Concept.} LLJ users should first reflect on the goal of the evaluation by thinking about what they are trying to evaluate and what a perfect scenario would look like (\textbf{Q.~\ref{q:system_what}-~\ref{q:system_desirable}}). LLJ users do not necessarily need to redefine the construct they are trying to evaluate; they can rely on existing literature. 
% \textcolor{red}{However make sure you are properly reusing other people definitions}
For instance, early work by \citet{mani_automatic_2001} characterizes summary quality along two primary dimensions: informativeness and acceptability and has served as a foundational framework for the development of evaluation metrics, including ROUGE \cite{lin_rouge_2004}.
LLJ users should also reflect on the potential ethical considerations associated with the evaluation process, including whether the test is appropriate in the first place~\citep{messick_test_1980} (\textbf{Q.~\ref{q:system_shouldyou}}). For example, work in machine psychology often applies psychological assessments developed for humans directly to LLMs and draws conclusions from their responses, rather than first redefining psychological constructs in ways that are appropriate for LLMs~\citep{löhn_machine_2024}.

% For example, \textcolor{red}{defining hate speech as XYZ might conflate communities reclaiming their slurs or sex work or maybe Melina political example parties?}
% For example, \melina{Melina example of data labeling in political parties + ou quelque chose sur le fait que les criteres en summarization exclus certains dialectes (there was a paper on this by alexandra/su lin?}

% For example, physiognomy---the idea that character or morality can be inferred from facial features---has been widely discredited~\citep{olapade_physiognomy_2025, emspak_facing_2017}, yet similar ideas persist in facial recognition systems that predict traits such as sexual orientation~\citep{emspak_facing_2017, sharpe_using_2017} or criminality~\citep{emspak_facing_2017, wu_responses_2017}. 
% These cases reflect conceptualization failures, as the attempt to define background concepts (e.g., criminality or homosexuality) as predictable facial features has been rejected by the community.

\noindent\textbf{\twemoji{1f4d0} Operationalization.} Once the systematized concept is clearly defined, LLJ users should specify how it is operationalized, namely through explicit evaluation criteria and their associated definitions~(\textbf{Q.~\ref{q:ope_criteria}-\ref{q:ope_how}}). They should draw on prior work by experts in evaluation to identify and define these criteria~(\textbf{Q.~\ref{q:ope_val}}), rather than introducing ad hoc definitions. If new criteria and definitions are proposed, LLJ users should explain how they were validated~(\textbf{Q.~\ref{q:ope_val}}).
Furthermore, it is important to consider the context in which the evaluation criteria were originally defined, and whether this context affects their applicability when reused in a different setting~(\textbf{Q.~\ref{q:ope_task}}). For instance, defining evaluation criteria for summarization is crucial, as different operationalizations capture different notions of what constitutes a good summary. Intrinsic metrics emphasize coherence and informativeness, often in tension, whereas extrinsic metrics evaluate usefulness in downstream tasks~\citep{mani_automatic_2001}.

%Limitations
\noindent\textbf{\twemoji{26a0} Limitations.} 
As the operationalization process provides a proxy for measuring the underlying systematized concept, certain dimensions may not be captured by the chosen criteria. It is therefore important to disclose the potential limitations of the operationalization, as these will affect subsequent interpretations of the system (\textbf{Q.~\ref{q:opelimit_miss}}).
For example, collecting harmless preferences has become central to safety alignment research~\citep{bai_training_2022}. While \citet{bai_training_2022} explicitly avoided defining ``harmlessness'' to encourage diversity, follow-up work shows that preference datasets still tend to reflect Western and majority perspectives~\citep{kirk_prism_2024}. 
Finally, LLJ users should also consider whose values are being used as the standard~\citep{messick_test_1980} and what implications this entails~(\textbf{Q.~\ref{q:opelimit_values}}).

The \mybox{YELLOW}{Evaluation Criteria Card} aims to support LLJ users through the operationalization process.
Once the context surrounding the use of the LLJ has been identified, a critical step in the evaluation process is to clearly define and document what is being evaluated. 
A body of work has shown that the ML and NLP communities have often struggled to clearly define the systematized concept under evaluation and to properly operationalize it~\citep{blodgett_stereotyping_2021}. 
% \khaoula{Maybe add more background on operationalization and operationalization failures in LLJs? \textit{In contrast to observations by \citep{chehbouni_neither_2025}, where LLJ users often introduce their own ad hoc definitions---such as when prompting an LLJ to evaluate fluency.}}
As such, we propose a set of guiding questions to help LLJ users better reflect on the goals of their evaluation. This section should be considered independently of the measurement tool chosen by researchers and practitioners. Both the definition of what is being evaluated and how this concept has been characterized in the literature should precede any decision about the indicators. In particular, study design should not be driven by the capabilities of the evaluation tool---in this case, the LLJ---but instead guided by the underlying goal of the evaluation.

\subsection{LLJ Pipeline}
\label{subsec:questions_llj_pipeline}

In the \mybox{GREEN}{LLJ Pipeline Card}, LLJ users should document their implementation, including the initial setup, the model(s) used, the final prompt(s), and the scoring function.

\noindent\textbf{\twemoji{2699} Set-Up.} LLJ users should report the functionality of the LLJ~(\textbf{Q.~\ref{q:pipeline_func}})---whether it is used for data-related tasks, training, or evaluation~\citep{li_llmsasjudges_2024}---as well as the type of configuration employed~(\textbf{Q.~\ref{q:pipeline_conf}}). This includes specifying whether they are using an LLJ alone, as part of an ensemble of models, or in combination with a human-in-the-loop evaluation process~\citep{li_llmsasjudges_2024}. We also ask LLJ users to indicate what is being evaluated~(\textbf{Q.~\ref{q:pipeline_what}}) (e.g., automated summaries in English, tweets for hate speech, etc.), along with the evaluation type~(\textbf{Q.~\ref{q:pipeline_type}}): pointwise, pairwise, or listwise, etc.---i.e., whether the assessment involves absolute or relative comparisons. For example, pairwise evaluation is particularly popular in safety alignment, when collecting LLM-based preferences to train a model~\citep{lee_rlaif_2024, bai_constitutional_2022, sun_salmon_2023}.

\noindent\textbf{\twemoji{1f916} Model.} LLJ users should report details of the final model selected for their evaluation~(\textbf{Q.~\ref{q:model_description}}), including the model family, version, size, date of access, and any sampling parameters used. They should also briefly justify their choice of model (e.g., state-of-the-art performance in the literature, open-source availability, or privileged access). In addition to documenting the LLJ, users should provide information about whether the model used is proprietary, if an open-source evaluation is provided and if the evaluation code is made publicly available~(\textbf{Q.~\ref{q:model_repro}}).
LLJ users should also report the costs and resources required for the final evaluation~(\textbf{Q.~\ref{q:model_budget}}). This may include API expenses for proprietary models or computational requirements for open-source models. 
% \khaoula{add example about how eval should be accessible? Not sure what would be the best example here. florian papers+example on how adding un budget peut etre incapacitant pour certains chercheurs dans le monde.}

\noindent\textbf{\twemoji{270d} Prompt.} LLJ users should provide examples of all final prompts they are using in their evaluation~(\textbf{Q.~\ref{q:prompt_structure}}). Best practices in prompt engineering usually recommend a prompt to have the following components: role, instructions, examples, chain-of-though or other prompt optimization strategy and input and output format~\citep{zhang_why_2025}. In the context of LLJs, the literature typically relies on an instruction—although some work has explored the use of personas for LLJ~\citep{wu_large_2023}---followed by an evaluation criterion and its definition (optional). Additional examples for few-shot learning or chain-of-thought can also be integrated, some users also request the LLJ to provide an explanation for the prediction. We refer the readers' to \citet{li_llmsasjudges_2024} for a more comprehensive account of potential LLJ design configurations. LLJ prompts can also contain reference material; for instance, in text summarization tasks, LLJ users may include a human-written gold summary as a benchmark for what an ``ideal'' summary should resemble, along with guidance on the expected output format. In all cases, LLJ users are expected to report their final prompt(s) to ensure reproducibility.

% Output: evaluation result is the primary output and can take multiple forms depending on evaluation type (ranking, binary preferred response, label, score on likert-scale, etc. with \citep{}). If an explanation was asked --- it can help interpret results despite no faithfulness, and feedback happens a lot in the context of model enhancement where LLJ are used to improve the training process = self-alignment examples

\noindent\textbf{\twemoji{1f3af} Score.} 
After scoring the cases, the LLJ typically produces an output that must be processed, either (1) to extract the final prediction or (2) to further transform the scores to account for LLM reliability issues and hallucinations.
We expect LLJ users to inform these extra steps in (\textbf{Q.~\ref{q:score_transfo}}) and to justify their post-processing choices (\textbf{Q.~\ref{q:score_justification}}). 
For example, when leveraging an LLJ for generating pairwise preferences for reinforcement learning with AI feedback, \citep{lee_rlaif_2024} extract the log probabilities of generating the tokens ``1'' and ``2'' and apply a softmax to obtain a preference distribution used as the final score.

The \mybox{GREEN}{LLJ Pipeline Card} aims to help LLJ users thoroughly document their final LLJ pipeline. Reproducibility is essential for valid and reliable evaluation, as it ensures that the process can be repeated and produces consistent results and conclusions~\citep{pineau_improving_2021}. However, in the context of LLJs, this assumption does not hold in practice. Many LLJ approaches rely on proprietary models that may be updated or deprecated without notice~\citep{liu_geval_2023}, making exact replication inherently difficult. Even when studies attempt to document their setups, key details are frequently underspecified---for instance, by referring to ``off-the-shelf'' models~\citep{bai_constitutional_2022} or to ``ChatGPT'' without specifying exact versions, omitting hyperparameters, or not fully reporting the content of prompts used in evaluation~\citep{chehbouni_neither_2025}. 
As a result, many LLJ pipelines are only partially reproducible. This card aims to address this under-specification by guiding LLJ users on the minimal information they should report to improve reproducibility and transparency. We structure it based on the conception of an LLJ system proposed by \citet{li_llmsasjudges_2024}.
% , as described in Section~\ref{sec:background}. 
% \khaoula{come back here after background}

\subsection{LLJ Evaluation}
\label{subsec:questions_llj_evaluation}

In the \mybox{PEACH}{LLJ Evaluation Card}, LLJ users should validate the evaluation outcomes as well as discuss the potential societal consequences of the evaluation.

\noindent\textbf{\twemoji{1f4ca} Evaluation.} LLJ users should report how they evaluated the scores derived from the LLJ outputs: the metrics, benchmarks, baselines, etc.~(\textbf{Q.~\ref{q:evaluation_how}}). Popular methods include correlation with human judgment, or performance on a downstream task when the LLJ scores are being used for training. For example, \citet{lee_rlaif_2024} assess the effectiveness of LLM-generated feedback for safety alignment by training a model on such feedback and then employing human evaluators to measure the resulting model’s win rate and harmlessness level.
% \melina{Melina best practices papers: When relying on human evaluation, best practices call for providing details about annotators’ backgrounds and recruitment, reporting inter-annotator agreement, and presenting disaggregated results.}
When evaluating LLJ scores, we also recommend presenting disaggregated results and statistical analysis if appropriate~(\textbf{Q.~\ref{q:evaluation_anal}}). 
We also ask LLJ users to document any tests conducted to assess validity~(\textbf{Q.~\ref{q:evaluation_val}}). In this context, correlations with human judgments and comparisons to baseline metrics are commonly used as evidence of convergent validity when validating a judge~\citep{chehbouni_neither_2025}. 
LLJ users should also disclose the limitations of their evaluation process~(\textbf{Q.~\ref{q:evaluation_limit}}), including anything excluded from the overall analysis. 
% \melina{Melina:For example, it is common in data annotation to discard cases because of low-quality annotation, these examples should be documented and disclosed in this step.} 
Finally, in some cases where the LLJ is intended for production in a dynamic environment, maintenance considerations should be taken into account~(\textbf{Q.~\ref{q:evaluation_dyna}}): how frequently should the LLJ be re-evaluated? 
% are there any retirement plans in place? 
% \textit{Data distribution shifts over time can adversely impact system performance in ways that may not be immediately apparent.}\citep{majumdar_red_2025}
% \khaoula{Explain why such a thing is needed with example of an LLJ for content moderation on social media.}   
% e some things excluded from analysis: \citet{scheuerman_datasets_2021}: \textit{Similarly, authors might outline refusals — data which they refused to collect, uses they refuse to condone, or opportunities for allowing data subjects or annotators to opt out — in their ethical statements [42].} \url{https://dl.acm.org/doi/10.1145/3406865.3419014}

\noindent\textbf{\twemoji{2696} Ethical Considerations.} LLJ users should reflect on the ethical considerations associated with the evaluation process and its outcomes. 
As such, we suggest that they begin by considering the potential consequences of a misleading evaluation~(\textbf{Q.~\ref{q:consequences_wrong}}), within the intended context of use~(\S~\ref{subsec:questions_context}). Taking inspiration from approaches such as impact statements~\citep{ashurst_guide_2020}, LLJ users should reflect on the benefits, risks, and potential uncertainties associated with the measurement process. This includes examining whether the evaluation process introduces or amplifies sources of invalidity related to bias, fairness, and distributive justice~\citep{messick_validity_1994}.  
% \khaoula{Add an example of having a wrong evaluation and what it means.}
%how does it compare
Furthermore, because anticipating unintended side effects of an evaluation is hard, we follow \citet{messick_test_1980} suggestion to approach this question comparatively, by contrasting the evaluation potential impacts with those of plausible alternatives~(\textbf{Q.~\ref{q:consequences_alter}}). For example, \citet{chehbouni_neither_2025} discuss how the field has progressively moved away from computationally efficient metrics (such as ROUGE~\citep{lin_rouge_2004}) toward the use of large LLMs, which come with substantially higher computational costs and, consequently, a potentially significant environmental footprint over time.
Finally, LLJ users are asked to reflect on the potential risks and harms that could arise from using the LLJ scores for action~(\textbf{Q.~\ref{q:consequences_use}}), either for decision-making, or or in subsequent applications. 
% \florian{In news summarization, the risk of focusing on relevance and coherence only is advancing the risks of systematic bias because of the way western country historically report news:\citep{moeller_regarding_2006}}

The {\mybox{PEACH}{LLJ Evaluation}} card aims to help LLJ users inform the final evaluation of the LLJ pipeline. 
This includes checking whether the operationalization is consistent with the final scores, whether the results are meaningful, and whether the claims made align with the scope of the process.
% As the assessment of measurement validity requires assessing if the scores truly capture the systematized concept as discussed in~\S~\ref{subsec:measurement_theory_background}.  
While measurement theory emphasizes that validity and reliability are not intrinsic properties of a test, but rather concern the interpretation of scores within a specific context~\citep{messick_validity_1994}, existing work on LLJs has often taken the opposite stance. In particular, LLJs are frequently assumed to be validated by prior studies and therefore directly transferable across tasks and settings, without requiring re-validation in new contexts~\citep{chehbouni_neither_2025, guerdan_validating_2025}. 
% For example, several studies reuse G-Eval for evaluation in substantially different applications and domains~\citep{}. 
As such, we encourage LLJ users to interpret the resulting scores in light of their limitations to avoid overgeneralization~\citep{messick_test_1980}: LLJ users may sometimes draw conclusions that exceed what the results can support, without properly validating that these results are generalizable~\citep{messick_test_1980}. \citet{crockett_ai_2026} refers to this phenomenon as the \textit{illusion of generalizability}, where the lack of grounding in a specific context enables cognitive scientists to overextend claims and assume their findings apply universally. This \textit{illusion of generalizability} is also prevalent in the NLP and ML literature, where it has contributed to the promotion of LLJ approaches presented as enabling universal evaluations, ``as one desires~\citep{fu_gptscore_2024}.''

\subsection{LLJ Prototyping}
\label{subsec:questions_pilot}

In the \mybox{PINK}{LLJ Prototyping Card}, LLJ users should document the prototyping process, including the study setup, tested data, LLJ design and validation, as well as any refinements made during this stage~\footnote{Note that although the LLJ Prototyping stage occurs earlier in the process, we describe it last, as it relies on concepts introduced in the LLJ Pipeline and LLJ Evaluation cards. We do not expect LLJ users to complete LLJ Pipeline and LLJ Evaluation cards for every configuration tested during the prototyping stage; however, the final LLJ pipeline should be thoroughly documented using these cards.}.
%Basically the idea behind this is in two phases (1) develop judge and optimize + (2) collect evidence for validity/reliability = and we do this on a subsample so it is more accessible
%To create subsample on s'inspire de red-teaming practices
%Merged with what we have seen in cybersecurity

% In this section, we introduce a five-step approach to the LLJ prototyping stage. 
% We therefore emphasize the importance of systematically recording all steps and procedures involved in this stage.

\noindent\textbf{\twemoji{23f3} Study Set-Up.} Based on the purpose of the evaluation~(\S~\ref{subsec:questions_context},\S~\ref{subsec:questions_evaluation_criteria}), LLJ users should first define the scope, timeline, expected cost of the LLJ prototyping phase, and determine whether specialized expertise is required~\citep{sabou_corpus_2014, peake_red_2021}~(\textbf{Q.~\ref{q:prototype_plan}}). For example, the smaller scope of this stage allows LLJ users to incorporate more diverse perspectives into the evaluation design process, such as by consulting the intended audience of the evaluation~\citep{singh_redteaming_2025, majumdar_red_2025}~(\S~\ref{subsec:questions_context}).
When determining the scope and depth of testing, \citet{peake_red_2021} recommend first identifying ``low-hanging fruit'': the most likely edge cases and known biases in LLJs that can be readily tested and are expected to yield meaningful insights if addressed. 
% \khaoula{For example, when using an LLJ as a safety classifier~\citep{} potential edge cases could include  =) write example from sorry safety paper}.

\noindent\textbf{\twemoji{1f3b2} Data.} LLJ users should construct the prototyping dataset which will serve as a baseline for optimizing the LLJ pipeline and assessing its validity.
To maximize its effectiveness, this dataset should include both a random sample of cases and challenging edge-case examples~\citep{klie_analyzing_2024, majumdar_red_2025}, as this increases the likelihood of uncovering potential weaknesses in the design early on~(\textbf{Q.~\ref{q:prototype_subsample}-\ref{q:prototype_edge}}). For example, difficult cases in data labeling might include data points with high disagreement between annotators, while difficult cases in text  summaries can refer to summaries with low compression rates.
% \florian{For example, difficult cases in text summarization might include summaries with high disagreement between annotators, or summaries with a high compression rate=any ref for low compression rate and disagreement in summarization?.} \khaoula{(et expliquer que edge case they need to justify based on their context}

\noindent\textbf{\twemoji{1f6e0} Design.} LLJ users should document and disclose the different model they used, their prompting strategies and configuration choices across all evaluated variants, in line with prior recommendations on reproducibility and reporting in research~\citep{baumann_large_2025}~(\textbf{Q.~\ref{q:prototype_tuning}}). Furthermore, LLJ users should also disclose, if possible, the results of this preliminary tuning process~(\textbf{Q.~\ref{q:prototype_results}}), to increase transparency about potential biases or reliability issues~\citep{baumann_large_2025}.
% Indeed, while it is common practice to perform prompt engineering~\citep{zhang_why_2025} before converging on a final LLJ pipeline~\khaoula{(ref to section)}, these design choices are rarely fully reported by LLJ users. \khaoula{I moved this in other section so just say that we suggest llj users fully document this process} 
% \khaoula{add example from llj literature of reporting?.} \khaoula{report results=of min and max of what you tried} 

% Doing so will enhance transparency, especially since testing different prompt formulations has been shown to improve robustness~\citep{baumann_large_2025}. \khaoula{add example from llj literature of reporting?.}

\noindent\textbf{\twemoji{1f9ea} Validation.} LLJ users should conduct a range of evaluations targeting different forms of validity and reliability to more comprehensively assess the system~(\textbf{Q.~\ref{q:prototype_val}-\ref{q:prototype_rel}}).
A useful way to approach this process is to focus on unacceptable behavior: after having defined the system’s desired outputs~(\S~\ref{subsec:questions_evaluation_criteria}), we encourage LLJ users to consider what the model should not do, as this ``negative space'' is often easier to articulate~\citep{majumdar_red_2025, singh_redteaming_2025}. This perspective can then be used to design targeted tests that explicitly probe for such undesirable behaviors. 
% \khaoula{Write example based on what i did in reproduciblity===For example, to test for \textit{face validity}, LLJ users can take advantage of the small size of the dataset to conduct manually inspections, while various test can be conducted for...}

\noindent\textbf{\twemoji{1f501} Refinement.} LLJ users should, after assessing validity, consider whether any refinement of the LLJ pipeline is required, or in some cases whether it is necessary to even modify the systematized concept~(\textbf{Q.~\ref{q:prototype_refine}}).
% (\khaoula{as seen in Figure}). \melina{add example from data labeling of refinement needed after pilot study == maybe we can put jackie's example here}.
A key question in this refinement process is when to stop~(\textbf{Q.~\ref{q:prototype_stop}}): determining what constitutes ``good enough'' performance is inherently task-dependent and varies across domains, task complexity, and evaluation goals. In some cases, even moderate or imperfect agreement scores may be acceptable if they are consistent with the state of the art or aligned with the intended use of the system. 
% \khaoula{Basically it is a little feedback loop of eval criteria, llj pipeline, llj eval}

The {\mybox{PINK}{LLJ Prototyping}} card is intended to support LLJ users in designing the final LLJ pipeline and LLJ evaluation protocol.
A common practice in data labeling, which we propose to adopt for LLJ use, is conducting a pilot study~\citep{klie_analyzing_2024,feng_acquiring_2009, vertanen_imagination_2011,sabou_corpus_2014}. Pilot studies are small-scale, initial experiments aimed at developing and refining the annotation process until it is sufficiently robust for full-scale deployment~\citep{klie_analyzing_2024}. During this phase, practitioners iteratively improve annotation guidelines, clarify task definitions, and adjust methodological choices~\citep{klie_analyzing_2024}. 
Introducing such a phase is particularly important in the context of LLJs, where evaluations can be costly, whether due to API usage or computational demands. A pilot study enables practitioners to identify and address issues early, before committing substantial resources to a flawed evaluation setup~\citep{klie_analyzing_2024}. During pilot studies, it is common to uncover limitations in the annotation guidelines, which may fail to capture certain phenomena, or be ambiguous or difficult for annotators to interpret~\citep{klie_analyzing_2024}. When such issues arise, it is appropriate to revisit the initial guideline design and iteratively refine and improve it before proceeding to full-scale evaluation~\citep{bareket_neural_2021}. This idea is closely aligned with practices in red-teaming~\citep{inie_summon_2025}, where systems are iteratively stress-tested to identify failures and unintended behaviors under challenging conditions. 
These benefits make a strong case for incorporating a pilot-study phase as a standard component of LLJ use. Since pilot studies traditionally involve human participants, we refer to the analogous process in LLJ workflows as the LLJ prototyping stage. 
Finally, while LLJ users typically already implement some of these practices---such as testing multiple models, prompt engineering, and stress-testing evaluators through prompt sensitivity or position bias checks~\citep{li_llmsasjudges_2024}---they often do so informally and without consistent documentation. With this card, we aim to formalize the process and improve transparency, while also reducing workload by concentrating the most intensive testing on a small, targeted subsample of cases.

% \section{LLJ Cards in Practice}
% \label{sec:lljcards_filled}

% We have made our framework available on Hugging Face Spaces~\footnote{https://huggingface.co/spaces}
% to facilitate access and allow users to contribute to a shared collection of LLJ cards. While we do not disclose the link to the shared spaces for anonymity purpose, Appendix~\ref{app:llj_huggingface} presents screenshot of that workspace. In Appendix~\ref{app:llj_cards_geval} we apply our framework to \citet{liu_geval_2023} seminal paper on LLJs for text summarization, which has since been widely reused in subsequent work, and showcases an example of gaps in LLJ documentation. 

% % Appendix~\ref{} presents example completed cards, while Appendix~\ref{} presents screenshots of the Hugging Face Space.

%  % \khaoula{Write how applying our sheet to these cases showed how some part were not documented -- underspecified reporting, and in our sheet we can compensate. Examples de plusieurs cas d'applications dans l'annexe --- en pratique on voit que nos fiches complete l'info manquante sur ces points (contexte/eval criteria) + alors qu'en parallele llj pipeline more documented, meme si ya pas tous les details et eval okay}

% \subsection*{LLJ Cards in Practice}
% \label{subsec:using_cards}

\section{Conclusion}
\label{sec:conclusion}

% Despite their widespread adoption, a growing body of work has highlighted important concerns with LLJ-based evaluations~\citep{chehbouni_neither_2025}. Nevertheless, their popularity reflects a clear need within the community: a scalable response to the growing evaluation challenges that the ML and NLP communities are facing~\citep{}. 
We have introduced a practical tool designed to support the valid and reliable use of LLJs in evaluation settings. Our methodological framework, \textbf{LLJ Cards}, aims to improve evaluation practices while enhancing transparency and reproducibility in the field. 
% \textbf{LLJ Cards} are inspired by established best practices for evaluation across the social sciences, computational social science, and machine learning and natural language processing. 
While we hope our framework can be applied to any LLJ implementation, it was primarily developed and illustrated through three applications: text summarization, safety alignment, and data labeling. Although these settings differ substantially in how LLJs are used and correspond to distinct communities of practice, they do not cover the full range of possible LLJ applications. 
Our framework is intended to be flexible, however we caution LLJ users about potential limitations in its generalizability across settings. 
% Appendix~\ref{ap:limitations} provides additional discussion of the limitations and ethical considerations of our work.
We believe there is strong promise in scalable automated evaluation when it is done rigorously---for instance, by enabling more realistic, interactive, and long-horizon evaluation pipelines that better reflect real-world usage. However, realizing this potential requires principled and careful use. We hope that our framework contributes to supporting such practice and encourages more responsible deployment of LLJ-based evaluation.

\section{Acknowledgements}

Funding support for project activities has been
partially provided by Canada CIFAR AI Chair,
NSERC Discovery grant and FRQNT fellowship. We
also express our gratitude to the Digital Research Alliance of Canada for their support in providing facilities for our evaluations.

% \begin{itemize}
%     \item We have seen how our cards answer known problems in LLJ use
%     \item We hope it will help with practices/transparency
%     \item Even being aware of this can help
%     \item Discuss main limitations 
%     \item Refer to Appendix with limitations/ethical considerations~\ref{ap:limitations}
%     \item Limitation method: best practices + everything observed viennent de 3 disciplines---on garantit pas l'exhaustivité de toutes les pratiques ou les gens utilisent llj
%     \item  NOT EXHAUSTIVE IT'S MINIMUM INFORMATION REQUIRED
%     \item WE TALK A LOT ABOUT HOW CONTEXTUAL APPROACH IS IMPORTANT SO SAME HERE NOT EVERYTHING IS GENERALIZABLE ACROSS ALL LLJ CONTEXT AS WE ONLY FOCUSED ON 3 TYPES OF APPLICATIONS, on a préféré aller dans la profondeur plutot que la largeur
%     \item We believe that there is promise for scalable automated evaluation, when done properly, as 
% \end{itemize}

\bibliographystyle{plainnat}
\bibliography{references_lljschecklist}

%%%%%%%%%%%%%%%%%%%%%%%%%%%%%%%%%%%%%%%%%%%%%%%%%%%%%%%%%%%%

% \appendix

% \section{Measurement Theory and LLM as Judges}
% \label{app:figure_measurement}

% % --------- Melina ----------

% \section{LLJ Cards Hugging Face}
% \label{app:llj_huggingface}

% This appendix present an overview of our various\textbf{ LLJ Cards} in HuggingFace Spaces.

% \label{app:llj_cards}
% \includepdf[
%     pages=-,
%     width=\textwidth,
%     height=\textheight,
%     keepaspectratio,
%     % pagecommand={
%     %     \section*{Appendix A: LLJ Cards Hugging Face}
%     % }
% ]{figures/LLJ1.pdf}

% \newpage
% \includepdf[
%     pages=-,
%     width=\textwidth,
%     height=\textheight,
%     keepaspectratio,
%     % pagecommand={
%     %     \section*{Appendix B: LLJ Cards Hugging Face}
%     % }
% ]{figures/LLJ2.pdf}

% \newpage
% \includepdf[
%     pages=-,
%     width=\textwidth,
%     height=\textheight,
%     keepaspectratio,
%     % pagecommand={
%     %     \section*{Appendix C: LLJ Cards Hugging Face}
%     % }
% ]{figures/LLJ11.pdf}

% \newpage
% \input{geval}

% \newpage
% \input{checklist}

\end{document}

%% file: llj_checklist_fixed.tex
\resetcounter

% ================= TWO COLUMNS =================

% \begin{multicols}{2}

% ===== LEFT COLUMN — BOX 1 =====
\begin{minipage}[t]{0.49\textwidth}
\fontsize{8}{10}\selectfont
% Reduce line spacing globally
\renewcommand{\baselinestretch}{0.9}  % Default is 1.0 (smaller number = tighter lines)
\setlength{\parskip}{0pt}             % Remove space between paragraphs
\setlength{\itemsep}{0pt}             % Remove space between list items

\setlength{\multicolsep}{0pt}     % Space between columns (top/bottom of multicols)
\setlength{\premulticols}{0pt}    % Space before multicols
\setlength{\postmulticols}{0pt}   % Space after multicols
% ================= COMPACT LIST SPACING =================
\setlist{
  itemsep=1pt,
  topsep=2pt,
  parsep=0pt,
  partopsep=0pt
}

% ================= SECTION BOX =================
% Fix spacing between tcolorboxes
\tcbset{
  section/.style={
    width=\columnwidth,
    colback=LIGHT,
    colframe=#1,
    colbacktitle=#1,
    coltitle=black,
    fonttitle=\bfseries,
    boxrule=0.5pt,
    arc=2pt,
    left=6pt,
    right=6pt,
    top=3pt,
    bottom=3pt,
    noparskip,
    before=\par\vspace{0pt},  
    after=\par\vspace{2pt}     
  }
}

\begin{tcolorbox}[section=BLUE, title=\ref{subsec:questions_context} Context]

\subtitle{\raisebox{-0.2em}{\scalebox{1.4}{\twemoji{1f30e}}}\hspace{0.55em}Application}
    \numitem{What is the domain of application?}\label{q:app_domain}
    \numitem{What are known risks in the environment?}\label{q:app_risks}

\subtitle{\raisebox{-0.2em}{\scalebox{1.4}{\twemoji{1f4dd}}}\hspace{0.5em}Task}
    \numitem{What is the task?}\label{q:task_what}
    \numitem{What are the known limitations of the task?}\label{q:task_limitations}
    \numitem{How does the task impact different populations?}\label{q:task_bias}

\subtitle{\raisebox{-0.2em}{\scalebox{1.4}{\twemoji{1f4c1}}}\hspace{0.5em}Cases}
    \numitem{What are the cases?}\label{q:data_info}
    \begin{multicols}{2} 
    \scriptsize
        \begin{attributes}
              \item Source
              \item Domain
              \item Scope
              \item Limitations
              \item Values
              \item Repurposing
        \end{attributes}
    \end{multicols}    
\subtitle{\raisebox{-0.2em}{\scalebox{1.4}{\twemoji{1f3e2}}}\hspace{0.3em} Stakeholders}
    \numitem{What is the context of use?}\label{q:llj_context}
    \numitem{Who is the intended audience of the evaluation?}\label{q:llj_audience}

    \numitem{What is the background of the LLJ users?}\label{q:llj_positionality}

\end{tcolorbox}
% \end{minipage}

% ===== LEFT COLUMN — BOX 3 =====
% \begin{tcolorbox}[section=PURPLE, title=Evaluation Criteria]
% \begin{minipage}[t]{0.49\textwidth}
\begin{tcolorbox}[section=YELLOW, title=\ref{subsec:questions_evaluation_criteria} Evaluation Criteria]\label{card:eval_criteria}

% \hfill \scalebox{1.2}{\twemoji{1f7e9}} \scalebox{1.2}{\twemoji{1f7e7}} \scalebox{1.2}{\twemoji{1f7e6}}]

\subtitle{\raisebox{-0.25em}{\scalebox{1.4}{\twemoji{1f4a1}}}\hspace{0.7em}Systematized Concept}
    \numitem{What are you trying to evaluate?}\label{q:system_what}
    \numitem{What constitutes desirable system behavior?}\label{q:system_desirable}
    \numitem{Should the concept be evaluated at all?}\label{q:system_shouldyou}
\subtitle{\raisebox{-0.2em}{\scalebox{1.4}{\twemoji{1f4d0}}}\hspace{0.7em}Operationalization}
    \numitem{What are the evaluation criteria?}\label{q:ope_criteria}
    \numitem{How were they selected?}\label{q:ope_how}
    \numitem{How were they validated?}\label{q:ope_val}
    \numitem{Are the criteria definitions task-dependent?}\label{q:ope_task}
        \begin{attributes}
          \item If so, have they been validated in this context?
        \end{attributes}     
\subtitle{\raisebox{-0.2em}{\scalebox{1.4}{\twemoji{26a0}}}\hspace{0.7em}Limitations}
\numitem{What do the criteria miss?}\label{q:opelimit_miss}
\numitem{Whose values do they embed?}\label{q:opelimit_values}
\end{tcolorbox}
% \end{minipage}
% ===== RIGHT COLUMN — BOX 5 =====
% \begin{minipage}[t]{0.49\textwidth}
\begin{tcolorbox}[section=PINK, title=\ref{subsec:questions_pilot} LLJ Prototyping]

\subtitle{\raisebox{-0.2em}{\scalebox{1.4}{\twemoji{1f4c5}}}\hspace{0.6em}Study Set-Up}
    \numitem{Planning:}\label{q:prototype_plan}
    \begin{multicols}{2} 
    \scriptsize
        \begin{attributes}
              \item Scope
              \item Time
              \item Cost
              \item Stakeholders
        \end{attributes}
    \end{multicols}  
   
\subtitle{\raisebox{-0.2em}{\scalebox{1.4}{\twemoji{1f522}}}\hspace{0.6em}Data}
    \numitem{How did you create your test set?}\label{q:prototype_subsample}
    \numitem{Did you include edge/hard cases?}\label{q:prototype_edge}
\subtitle{\raisebox{-0.2em}{\scalebox{1.4}{\twemoji{1f3a8}}}\hspace{0.6em}Design}
    \numitem{Which models/prompts/config were tested?}\label{q:prototype_tuning}
    \numitem{Did you report the results of your tuning phase?}\label{q:prototype_results}
\subtitle{\raisebox{-0.2em}{\scalebox{1.4}{\twemoji{2705}}}\hspace{0.6em}Validation}
    \numitem{Have you tested your LLJ for validity?}\label{q:prototype_val}
    \numitem{Have you tested your LLJ for reliability?}\label{q:prototype_rel}
\subtitle{\raisebox{-0.2em}{\scalebox{1.4}{\twemoji{1f501}}}\hspace{0.6em}Refinement}
    \numitem{Have you made any changes to the evaluation criteria following the LLJ tuning?}\label{q:prototype_refine}
    \numitem{What was the stopping criterion?}\label{q:prototype_stop}

\end{tcolorbox}
\end{minipage}
% ===== LEFT COLUMN — BOX 2 =====
\begin{minipage}[t]{0.49\textwidth}
\fontsize{8}{10}\selectfont
% Reduce line spacing globally
\renewcommand{\baselinestretch}{0.9}  % Default is 1.0 (smaller number = tighter lines)
\setlength{\parskip}{0pt}             % Remove space between paragraphs
\setlength{\itemsep}{0pt}             % Remove space between list items

\setlength{\multicolsep}{0pt}     % Space between columns (top/bottom of multicols)
\setlength{\premulticols}{0pt}    % Space before multicols
\setlength{\postmulticols}{0pt}   % Space after multicols
% ================= COMPACT LIST SPACING =================
\setlist{
  itemsep=1pt,
  topsep=2pt,
  parsep=0pt,
  partopsep=0pt
}

% ================= SECTION BOX =================
% Fix spacing between tcolorboxes
\tcbset{
  section/.style={
    width=\columnwidth,
    colback=LIGHT,
    colframe=#1,
    colbacktitle=#1,
    coltitle=black,
    fonttitle=\bfseries,
    boxrule=0.5pt,
    arc=2pt,
    left=6pt,
    right=6pt,
    top=3pt,
    bottom=3pt,
    noparskip,
    before=\par\vspace{0pt},  
    after=\par\vspace{2pt}     
  }
}

\begin{tcolorbox}[section=GREEN, title=\ref{subsec:questions_llj_pipeline} LLJ Pipeline]

\subtitle{\raisebox{-0.2em}{\scalebox{1.4}{\twemoji{2699}}}\hspace{0.5em}Set-Up}
    \numitem{Functionality:             \opt{Data}\opt{Training}\opt{Eval}\opt{Other}}\label{q:pipeline_func}    
    \numitem{Configuration: \opt{LLJ}\opt{LLJ Jury}\opt{HITL}\opt{Other}}\label{q:pipeline_conf}
    \numitem{What is being evaluated?}\label{q:pipeline_what}
    \numitem{What is the evaluation type?}\label{q:pipeline_type}

\subtitle{\raisebox{-0.2em}{\scalebox{1.4}{\twemoji{1f916}}}\hspace{0.5em}Model}
    \numitem{Model Description:}\label{q:model_description} 
        \begin{multicols}{2}
        \scriptsize
            \begin{attributes}
              \item Family
              \item Version
              \item Size
              \item Date accessed
              \item Hyperparameters
              \item Justification 
            \end{attributes}
        \end{multicols}
    \numitem{Reproducibility:}\label{q:model_repro} 
      \begin{attributes}
          \item Proprietary (Y/N)?
          \item If yes, do you offer an open-source alternative?
          \item Is the LLJ code publicly available?
        \end{attributes}
    \numitem{Budget:}\label{q:model_budget} 
        \begin{attributes}
              \item What is the evaluation’s API cost?
              \item What is the evaluation’s computational cost?
        \end{attributes}

\subtitle{\raisebox{-0.2em}{\scalebox{1.4}{\twemoji{270d}}}\hspace{0.5em}Prompt}
\numitem{Prompt Structure:}\label{q:prompt_structure} 
    \begin{multicols}{2}
        \scriptsize
            \begin{attributes}
              \item Role
              \item Instructions
              \item Evaluation Criteria
              \item Criteria Definition
              \item Optimization
              \item References
              \item Input 
              \item Output
            \end{attributes}
    \end{multicols}

\subtitle{\raisebox{-0.2em}{\scalebox{1.4}{\twemoji{1f3af}}}\hspace{0.5em}Score}
    \numitem{How is the final score computed?}\label{q:score_transfo}
    \numitem{Why report the evaluation results this way?}\label{q:score_justification} 

\end{tcolorbox}
% \end{minipage}
% ===== RIGHT COLUMN — BOX 4 =====
% \begin{minipage}[t]{0.49\textwidth}
\begin{tcolorbox}[section=PEACH, title=\ref{subsec:questions_llj_evaluation} LLJ Evaluation]

\subtitle{\raisebox{-0.2em}{\scalebox{1.4}{\twemoji{1f4ca}}}\hspace{0.55em}Evaluation}
    \numitem{How are you evaluating the LLJ?}\label{q:evaluation_how}
        \begin{attributes}
          \item What metrics/benchmarks are you using?
          % \item Are you conducting human evaluation?
          \item What are the baselines, if any?
        \end{attributes}
    \numitem{Are you presenting non-aggregated results and additional analysis?} \label{q:evaluation_anal}
    \numitem{Are you assessing the validity of the scores?}\label{q:evaluation_val} 
    \numitem{Are you disclosing the limitations of the evaluation process?}\label{q:evaluation_limit} 
    % \numitem{Were any elements excluded from the evaluation process or evaluation results? If so, why?} 
    \numitem{Is this a dynamic evaluation?}\label{q:evaluation_dyna} 
    \begin{attributes}
        \item If so, what are the plans for maintaining or retiring the system?
    \end{attributes}

\subtitle{\raisebox{-0.2em}{\scalebox{1.4}{\twemoji{2696}}}\hspace{0.55em}Ethical Considerations}
    \numitem{What harms could arise from incorrect evaluation results?}\label{q:consequences_wrong}
    \numitem{How do the social impacts of this evaluation compare to alternative approaches?}\label{q:consequences_alter} 
    \numitem{What harms or risks could arise from using these evaluation results to guide decision-making or as a basis for further work?}\label{q:consequences_use}
\end{tcolorbox}
\end{minipage}